\documentclass[lettersize,journal]{IEEEtran}
\usepackage{amsmath,amsfonts}
\usepackage{algorithmic}
\usepackage{algorithm}
\usepackage{array}
\usepackage[caption=false,font=normalsize,labelfont=sf,textfont=sf]{subfig}
\usepackage{textcomp}
\usepackage{stfloats}
\usepackage{url}
\usepackage{verbatim}
\usepackage{graphicx}
\usepackage{cite}
\usepackage{hyperref}
\usepackage{caption}
\usepackage{adjustbox}
\usepackage{multirow}
\usepackage[table,xcdraw]{xcolor}
\usepackage{subfig}
\begin{document}

\title{Dual Adversarial Fine-tuning for Enhancing Robustness of Large Vision Language Model}

\author{Sibo Wang,~\IEEEmembership{Student Member,~IEEE,}
        Jie Zhang,~\IEEEmembership{Member,~IEEE,}
        Shiguang Shan,~\IEEEmembership{Fellow,~IEEE,}
        Xilin Chen,~\IEEEmembership{Fellow,~IEEE,}
        Wen Gao,~\IEEEmembership{Fellow,~IEEE}}

\markboth{Journal of \LaTeX\ Class Files,~Vol.~14, No.~8, August~2021}%
{Shell \MakeLowercase{\textit{et al.}}: A Sample Article Using IEEEtran.cls for IEEE Journals}


\maketitle

\begin{abstract}
While Large Vision-Language Models (LVLMs), represented by LLaVA and GPT-4V, have demonstrated remarkable capabilities, their visual inputs remain vulnerable to adversarial attacks, posing significant security risks. 
Existing defense methods predominantly target single-task scenarios (e.g., zero-shot classification) and consequently lack generalizability across various multimodal tasks. 
To address this limitation, we propose a dual adversarial fine-tuning framework that jointly optimizes visual and semantic supervision signals from two modalities, enhancing model robustness while generalizing across multiple downstream tasks. 
The proposed framework comprises two core components, i.e., \textbf{Visual} supervision branch and \textbf{Semantic} supervision branch. 
The former branch leverages features from clean images, extracted via a frozen original vision encoder, to guide adversarial robustness  while the latter incorporates caption-image alignment as a contextual signal to preserve semantic coherence under attack. Moreover, our method achieves cross-task robustness by simply replacing the CLIP vision encoder in the original model, with no need of separate task-specific retraining or architecture modifications.
Extensive experiments demonstrate that our approach outperforms the state-of-the-art method in adversarial robustness evaluation across zero-shot classification, image captioning, and visual question answering (VQA) tasks.
\end{abstract}


\section{Introduction}
\label{intro}
\IEEEPARstart{R}{ecent} advances in Large Vision-Language Models (LVLMs) have achieved significant progress in understanding and reasoning, with performance occasionally rivaling or surpassing human capabilities in specific tasks \cite{liu2024improved,liu2023visual,zhu2023minigpt}. 
These models now serve as foundational tools for diverse applications, including visual question answering, image captioning, and cross-modal retrieval \cite{chen2022visualgpt,jia2021scaling,wang2025adaptingmllm}. 
A significant milestone in this area is visual-language alignment frameworks such as CLIP \cite{radford2021learning}, which leverages contrastive learning to map text and image into a unified feature space. 
By minimizing the distance between semantically related pairs across modalities, CLIP achieves remarkable zero-shot generalization, enabling seamless adaptation to numerous downstream tasks without task-specific fine-tuning. 
Furthermore, CLIP serves as a foundational pillar for various Large Vision-Language Models (LVLMs). 
For example, LLaVA \cite{liu2023visual,lin2026moellava} integrates CLIP's vision encoder with the large language model Vicuna \cite{chiang2023vicuna}, enabling joint processing of visual and textual data and yielding exceptional performance on diverse downstream tasks, including image captioning, visual question answering (VQA), and text-image retrieval.

\begin{figure}[t]
    \centering
    \subfloat[Robust zero-shot classification accuracy on different datasets.\label{fig:fig1a}]{
        \includegraphics[width=0.95\linewidth]{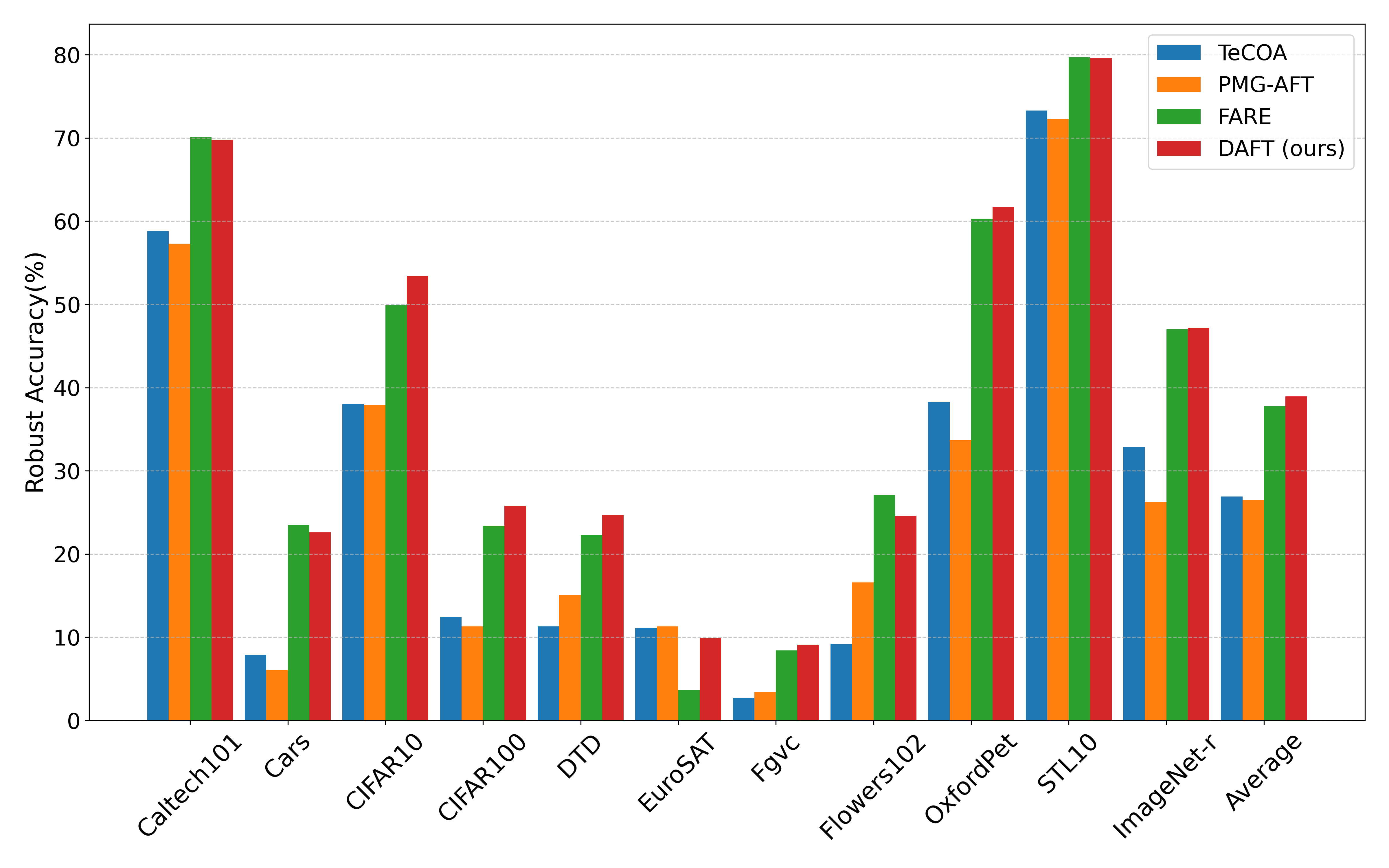}
    }

    \vspace{0.4em}

    \subfloat[Robust performance on captioning and VQA tasks.\label{fig:fig1b}]{
        \includegraphics[width=0.95\linewidth]{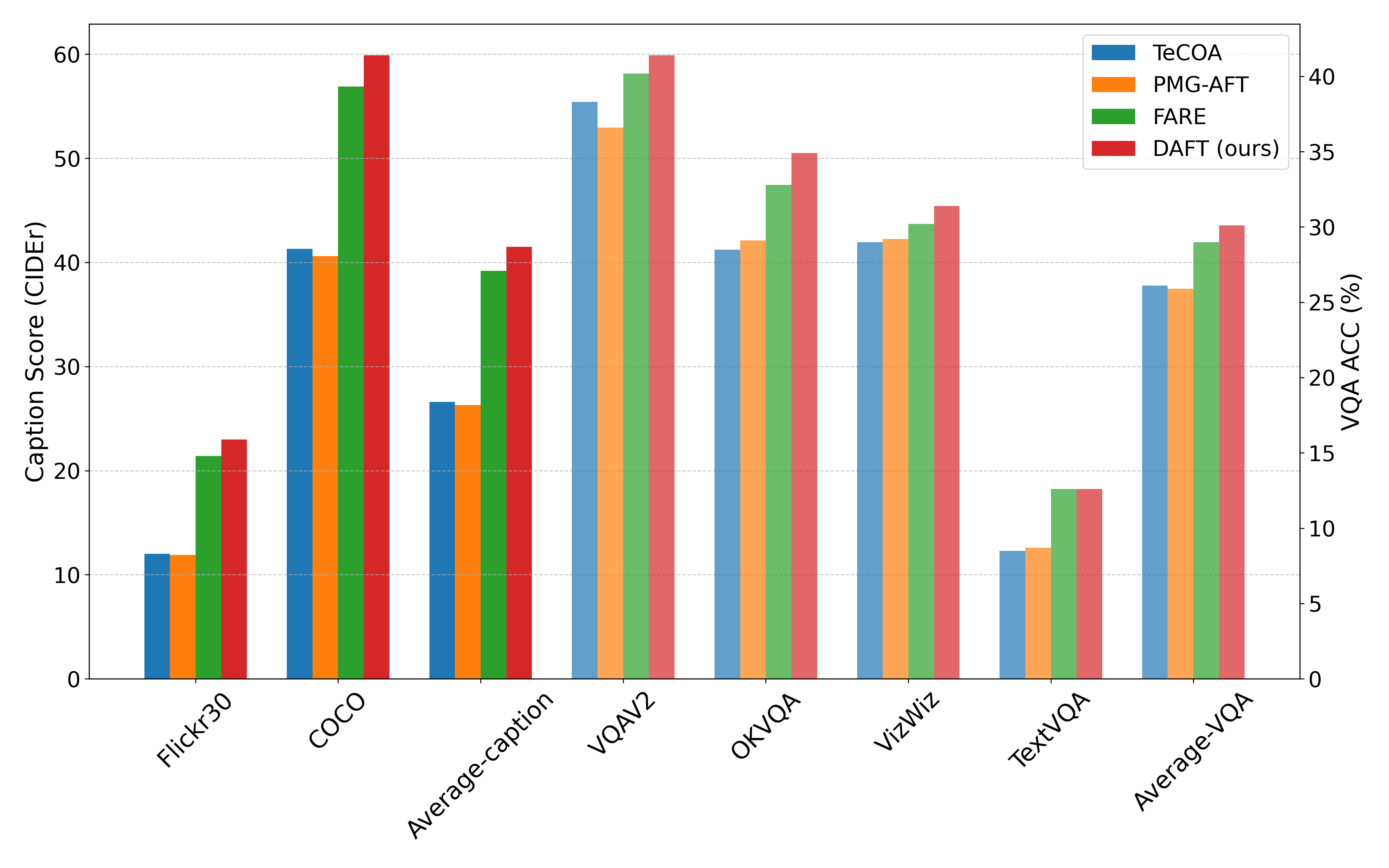}
    }

    \caption{Results under perturbation budget $\epsilon=2/255$.
    (a) Adversarial accuracy of CLIP with different fine-tuning methods on zero-shot classification.
    (b) Robust captioning and VQA performance after replacing the original vision encoder of LLaVA with different CLIP vision encoders.}
    \label{fig:figure1}
\end{figure}

Despite their capabilities, LVLMs remain inherently vulnerable to visual adversarial attacks, potentially leading to erroneous outputs, even in closed-source models \cite{zhao2023evaluating,schlarmann2023adversarial,carlini2023aligned,dong2023robust}.
These attacks introduce human-imperceptible noise to visual inputs, significantly degrading the model's performance across tasks \cite{carlini2017towards,goodfellow2014explaining,madry2017towards}. 
For example, previous studies have found that subtle perturbations can force LVLMs to generate predefined caption \cite{schlarmann2023adversarial}, or produce incorrect answers in visual question answering (VQA), highlighting systemic vulnerabilities in multimodal perception \cite{zou2023universal}.
Since an increasing number of LVLMs are deployed in security-sensitive applications, improving their adversarial robustness has become a critical priority.

\begin{figure*}[t]
\centering
\includegraphics[width=0.95\textwidth]{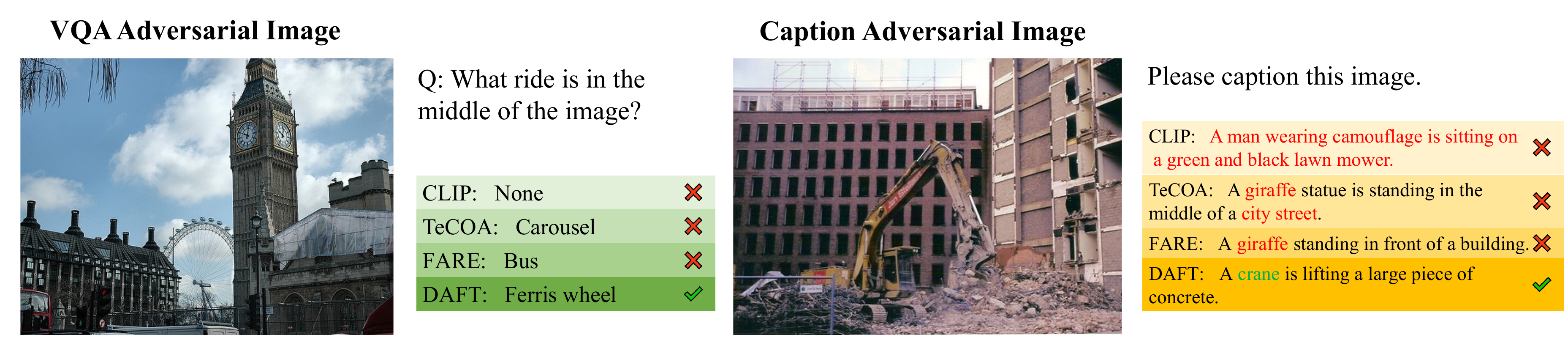}
\caption{Examples of robust evaluation on VQA (left) and image caption (right) tasks. 
We perform untargeted $\mathcal{L}_\infty$ attack with perturbation budget $\epsilon$ = 2/255 on LLaVA 1.5-7B and replace the original model with CLIP fine-tuned using different methods for inference. 
In VQA task, the responses from the original and TeCOA\cite{mao2022understanding} trained models are almost unrelated to the image-question pair. 
The FARE\cite{schlarmann2024robust} trained model responds the element ``bus" from the image, but it is not the correct answer. 
Our DAFT model, however, provides the most accurate answer ``Ferris wheel".
In image captioning, both the FARE\cite{schlarmann2024robust} and TeCoA\cite{mao2022understanding} mistakenly recognize the ``crane" as a ``giraffe", while our method accurately provides the caption for this image.}
\label{fig:intro_ex}
\end{figure*}

To mitigate security risks from adversarial examples, researchers have developed various defense strategies, among which adversarial training is regarded as one of the most effective defense mechanisms \cite{madry2017towards,pang2020bag,shafahi2019adversarial,zhang2019theoretically}. 
Its core idea is to dynamically generate adversarial examples during the training process to enhance the model’s robustness against adversarial attacks. 
However, in the context of large vision-language models (LVLMs), conducting adversarial training from scratch is impractical due to the excessive computational cost and data requirements. 
To address these challenges, recent work focuses on adversarial fine-tuning, which involves leveraging adversarial examples to fine-tune the vision encoder in LVLMs (e.g., CLIP’s vision encoder) and subsequently integrating it into LVLM to enhance overall robustness \cite{mao2022understanding,schlarmann2024robust,wang2024pre}.
Among existing studies, TeCoA \cite{mao2022understanding}, as a representative work, performs supervised adversarial fine-tuning on ImageNet \cite{deng2009imagenet}, combining adversarial image perturbations with category-aligned text prompts to improve CLIP’s zero-shot classification robustness. 
In contrast, FARE \cite{schlarmann2024robust} adopts unsupervised adversarial fine-tuning, further enhancing the robustness generalization of CLIP’s vision encoder without labeled data.
Critically, FARE allow LVLMs to inherit adversarial resilience simply by replacing their original vision encoder, eliminating the need for full-model retraining while improving adversarial robustness across various downstream tasks like image caption and visual question answering (VQA).

Despite advancements in adversarial robustness for LVLMs, two critical challenges remain:
\textbf{1) Category-Aligned Text Supervision Suffers from Overfitting and Poor Generalization.}
Existing methods usually rely on fixed text embeddings (e.g., ImageNet-1K class labels) for supervision, which optimizes robustness within predefined categories but neglects the impact of unseen classes. This narrow focus can lead to distortions in feature representations when models encounter tasks outside the training distribution.
Moreover, class labels lack the semantic richness for complex multimodal tasks like image captioning and visual question answering (VQA), limiting their applicability to classification-centric scenarios.
Consequently, the robustness improvements from label-supervised methods like TeCoA \cite{mao2022understanding} remain confined to classification tasks.
As shown in Fig. \ref{fig:figure1}, label-supervised methods (e.g., TeCoA \cite{mao2022understanding}, PMG-AFT \cite{wang2024pre}) exhibit limited robustness, in zero-shot classification tasks on unseen datasets and fail to generalize to vision-language tasks like VQA.
\textbf{2) Robustness Gains are Limited from Unsupervised Adversarial Fine-Tuning.}
Although unsupervised adversarial fine-tuning like FARE \cite{schlarmann2024robust} improves robust generalization to some extent, its overall adversarial robustness remains limited. 
The most advanced adversarial attacks on LVLMs primarily exploit vulnerabilities in image-text alignment to force mispredictions \cite{schlarmann2023adversarial,zhang2022towards,wang2025exploringtransferability}. 
Existing adversarial fine-tuning methods, however, focus narrowly on minimizing the feature discrepancies between clean and adversarial examples, making them ineffective against more sophisticated adversarial attacks that specifically target multimodal alignment mechanisms.
As shown in Fig. \ref{fig:fig1a}, unsupervised adversarial fine-tuning methods such as FARE \cite{schlarmann2024robust} exhibit strong robust generalization in zero-shot classification tasks. 
However, when facing more advanced attacks targeting large language vision models (as illustrated in Fig. \ref{fig:fig1b}), they falter, highlighting a critical gap in defense coverage compared to our method.
Fig.\ref{fig:intro_ex} further illustrates this limitation through concrete examples: while baseline methods struggle under adversarial conditions for vision-language tasks like VQA and image captioning, our method maintains consistent performance, underscoring its superior robustness.

Inspired by recent studies highlighting the critical role of captions in LVLM training and robust vision-language adaptation \cite{sun2024descriptive,saito2023prefix,waseda2025quality}, which demonstrate that rich textual captions surpass simplistic template prompts (e.g., “a photo of [class]”) in semantic expressiveness, we propose Dual Adversarial Fine-Tuning method (DAFT) to enhance the cross-task robustness in LVLMs through two synergistic branches. 
1) Semantic supervision branch: Replacing traditional cross-entropy loss with a caption-guided contrastive loss. 
This aligns adversarial examples with their original descriptive captions while distancing them from unrelated captions in the batch. 
2) Visual supervision branch: Preserving pre-trained knowledge by minimizing the feature distance between adversarial and clean examples via a frozen original vision encoder, thereby mitigating text-modality-induced overfitting.
We evaluate DAFT extensively on zero-shot classification, image captioning, and VQA across multiple datasets. 
Experimental results demonstrate that DAFT outperforms the state-of-the-art method in adversarial robustness for all tasks, achieving superior generalization by unifying visual and semantic alignment under attack conditions.

Our main contributions are summarized as follows:
\begin{itemize}
    \item We propose the DAFT, a dual adversarial fine-tuning framework that jointly optimizes visual and semantic supervision signals. It effectively mitigates overfitting while enhancing cross-task robustness in large vision-language models.
    \item To our knowledge, DAFT is the first framework to integrate descriptive captions (instead of class labels) as semantic supervision for adversarial traning. By leveraging contrastive learning to align adversarial examples with their original captions, we demonstrate significant robustness improvements in multimodal tasks.
    \item Extensive experiments demonstrate that DAFT consistently outperforms the state-of-the-art method in terms of adversarial robustness on zero-shot classification, image captioning, and VQA tasks.
\end{itemize}

\section{Related Work}

\noindent\textbf{Large Vision-Language Models.} 
Large Vision-Language Models (LVLMs) are the product of the convergence of natural language processing and computer vision. 
Compared to traditional large language models (LLMs), LVLMs are capable of handling both visual and textual modalities simultaneously, offering broader application scenarios ranging from visual question answering to image-grounded dialogue and complex reasoning tasks. 
LVLMs, such as LLaVA-1.5 \cite{liu2024improved,lin2026moellava} , BLIP-2 \cite{li2023blip} , MiniGPT-4 \cite{zhu2023minigpt}, Otter \cite{li2023mimic}, and InternLM-XComposer \cite{zhang2023internlm} , have demonstrated the ability to understand multimodal information and complete complex reasoning tasks. 
These models typically use a pre-trained vision encoder, such as CLIP, to transform images into vector representations (embeddings space), which are then aligned with text embeddings through various techniques. 
Some commercial LVLMs, such as Gemini \cite{team2023gemini} and GPT-4V\cite{achiam2023gpt}, achieve human-expert performance in certain domains. 
While most prior works have focused on enhancing the reasoning capabilities of LVLMs, our research shifts attention toward improving their robustness against adversarial attacks and evaluating the generalizability of this robustness across diverse downstream tasks.

\noindent\textbf{Adversarial Robustness.}
Deep neural networks have been found to be vulnerable to adversarial attacks \cite{athalye2018obfuscated,carlini2017towards,dong2018boosting,goodfellow2014explaining,croce2020reliable}, where carefully crafted imperceptible noise is added to the original images, causing the model to misclassify. 
To enhance the robustness of neural networks against adversarial examples, a series of defense strategies have been proposed. 
Among them, adversarial training \cite{madry2017towards,shafahi2019adversarial,wu2020adversarial,zhang2019theoretically} is considered the most effective defense mechanism. 
It dynamically generates adversarial examples during the training process and uses them to train the model, thereby improving the model's defense capabilities on the task it is trained for. 
With the rise of large vision-language models (LVLMs), their robustness has gradually become a focus of research, and numerous attack algorithms targeting them have emerged \cite{inkawhich2023adversarial,zhang2022towards,lu2023set,zhao2023evaluating,wang2025exploringtransferability,wang2026revisiting}. 
Since most vision-language models use CLIP's vision encoder and its variants as components, existing defense methods generally focus on adversarial fine-tuning of CLIP's vision encoder.

\noindent\textbf{Transferability of Robustness.}
Beyond improving robustness on training tasks, recent work begins to explore how adversarial robustness transfers to unseen tasks, such as zero-shot classification, captioning, and visual question answering. 
TeCoA \cite{mao2022understanding} introduces category text as labels, replacing traditional one-hot supervision, and achieves better robustness in zero-shot classification tasks. 
PMG-AFT \cite{wang2024pre} further introduces an auxiliary branch to minimize the distance between adversarial outputs of the target and pre-trained models, thus improving zero-shot adversarial robustness and mitigating overfitting. 
FARE \cite{schlarmann2024robust} employs unsupervised adversarial fine-tuning to bring robustness not only to classification but also to captioning and VQA. 
However, existing methods either tend to overfit to specific datasets or fail to fully capture the rich semantic relationships between images and texts. 
Our method introduces descriptive captions as semantic supervision and combines it with visual supervision, effectively addressing existing issues and enhancing the robustness transferability of large vision-language models.

\section{Methodology}
In this section, we introduce our DAFT framework. 
In Section \ref{Preliminaries}, we provide the background on adversarial attacks, adversarial training, and adversarial robustness generalization. 
Section \ref{sec:method detail} provides a detailed introduction to our adversarial fine-tuning method, including its components and loss functions.

\subsection{Preliminaries and Problem Setup}
\label{Preliminaries}
Since most current LVLMs use CLIP as the backbone, our work mainly focuses on adversarial fine-tuning of CLIP's vision encoder. 
It is worth noting that our method is also applicable to models with other backbones.
For the sake of formalization, we also take the CLIP model as our target model.
Let $F_{\theta}(\cdot)$ represent the CLIP image encoder parameterized by $\theta$ and $T_{\phi}(\cdot)$ represent the CLIP text encoder parameterized by $\phi$.
Given an input image $x$ and a textual input denoted as $t$, the model will provide an image representation $F_{\theta}(x)$ and a text representation $T_{\phi}(t)$ in a unified embedding space.
The text $t$ can either be the classification label of the image, such as ``a photo of [class]," or a description about the image. 
For the classification task, CLIP calculates the cosine similarity between the image embedding and the embeddings of each category label, then selects the category with the highest similarity as the classification result, i.e., $\arg\max_{k=1,\dots,K} \cos(F_{\theta}(x), T_{\phi}(t_k))$, where $t_k$ represents the text prompt for the $k$-th category label.

\begin{figure*}[t]
\centering
\includegraphics[width=0.98\textwidth]{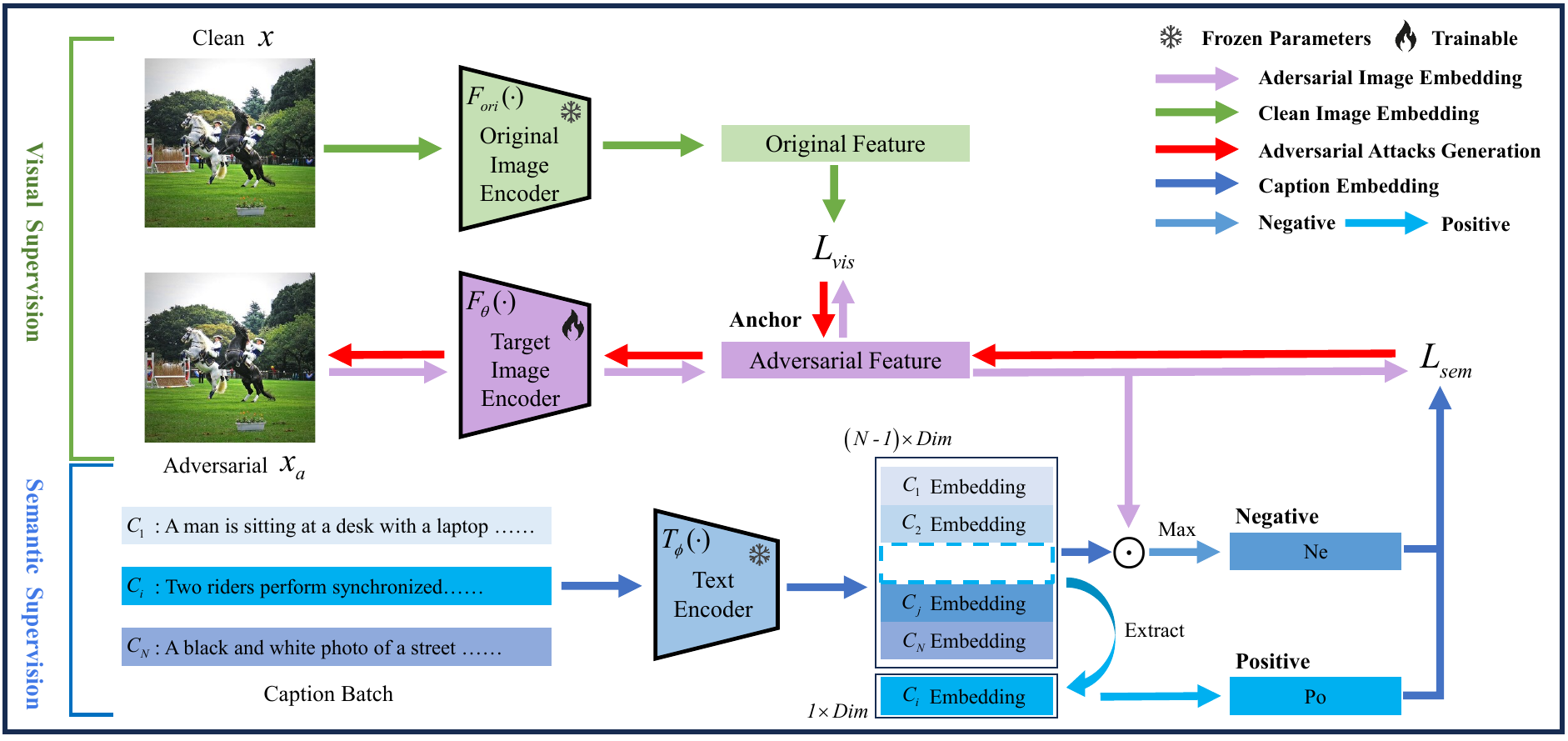}
\caption{The pipeline of DAFT. 
Visual supervision branch maximizes the similarity between adversarial feature on the target model and clean feature on the original model using $\mathcal{L}_\text{vis}$, ensuring generalization. 
Semantic supervision branch first uses the text encoder of the pre-trained CLIP model to encode a batch of $N$ captions, and generates positive and negative samples based on the relationship between the embeddings of these captions and the adversarial feature. 
The caption corresponding to the adversarial sample is considered the positive sample $Po$, while the caption with the highest similarity to the adversarial sample feature among the remaining $N-1$ captions is selected as the negative sample $Ne$.  
Furthermore, contrastive loss $\mathcal{L}_\text{sem}$ is used ensuring stronger robustness.
Only the image encoder of the target model can be trained and the adversarial examples generation alternates with parameters updating. 
$\odot$ means matrix inner product.}
\label{fig:pipeline}
\end{figure*}

\noindent\textbf{Adversarial Attacks.} 
Adversarial attacks typically refer to adding an imperceptible, optimizable perturbation to a clean image, misleading the model to produce an incorrect prediction or answer.
A well-known white-box attack method is Projected Gradient Descent (PGD) \cite{madry2017towards}, which uses multi-step gradient ascent steps to maximize the loss while projecting intermediate perturbation to the specified region constrained by the p-norm.
The implementation of PGD varies depending on the task or the attack objective. 
For the classification task, given the one-hot label $y$ , we first define a loss function:
\begin{equation}
\begin{aligned}
\label{loss-cls}
\mathcal{L}_{\text{adv}}=\mathcal{L}_{\text{ce}}([\text{cos}(F_{\theta}(x), T_{\phi}(t_1)), \dots, \text{cos}(F_{\theta}(x), T_{\phi}(t_k))],y),
\end{aligned}
\end{equation}
where $\mathcal{L}_{\text{ce}}$ denotes the cross-entropy loss.
For other more general scenarios, the loss function can be changed to 
\begin{equation}
\begin{aligned}
\label{loss-cls}
\mathcal{L}_{\text{adv}}=\mathcal{L}_{\text{distance}}(F_{\theta}(x),target),
\end{aligned}
\end{equation}
where $\mathcal{L}_\text{distance}$ represents a certain metric such as $\mathcal{L}_2$ distance, 
and $target$ represents objective such as the embedding of the original image or the corresponding text representation.
Different attack objectives can be achieved by modifying the implementation of the adversarial loss function.

Then, the adversarial perturbation is iteratively optimized to maximize this loss within a bounded perturbation region, typically using the following update rule:
\begin{equation}
\begin{aligned}
\label{pgd-cls}
x^{t+1} = \text{Proj}_{\epsilon}^p(x^t + \alpha \cdot \text{sign}(\nabla_x \mathcal{L}_\text{adv})),
\end{aligned}
\end{equation}
where $x^0 = x+\varepsilon_0$, $\varepsilon_0 \sim U[0, \varepsilon]$,  \( \text{Proj}_{\epsilon} ^p \) denotes the projection to \( \epsilon \)-ball under p-norm around the original image, \( \alpha \) is the step size, \( \nabla_x \mathcal{L}_\text{adv} \) is the gradient of the adversarial loss with respect to $x$.

\noindent\textbf{Adversarial Fine-Tuning.} Adversarial training formulates the training process into a min-max problem and can be described as:
\begin{equation}
\begin{aligned}
\label{minmax game}
\min_\theta \mathbb{E}_{x, y \sim P_D} \left[ \max_{x_a \in B(x,\varepsilon)} \mathcal{L}_\text{adv} \left( F_{\theta}(x_a), y \right) \right],
\end{aligned}
\end{equation}
where $F_{\theta}$ denotes the model parameterized by $\theta$, $\mathcal{L}_\text{adv}$ is the adversarial loss function mentioned above (e.g., cross-entropy) and $x_a$ is the adversarial example.
The inner maximization problem aims to find a stronger adversarial example, while the outer minimization problem optimizes the model parameters so that the model $F_{\theta}$ still makes correct predictions on the adversarial examples.
Since fine-tuning is generally used to adapt CLIP to downstream tasks, such an objective function \eqref{minmax game} can also be applied to the fine-tuning of CLIP towards robustness, and we refer to it as adversarial fine-tuning.

\noindent\textbf{Adversarial Robustness Generalization.} 
It is well known that LVLMs have excellent cross-task generalization capabilities. 
Whether the outstanding generalization ability of LVLMs across different tasks and datasets can remain consistent in the face of adversarial attacks is a question worth investigating. 
Adversarial fine-tuning a model for each task separately is an unscalable and costly approach. 
Therefore, we fine-tune the CLIP vision encoder only once and directly transfer and replace the original encoder of the LVLM (e.g., LLaVA \cite{liu2024improved}) with this encoder. 
We then test the adversarial test accuracy of LLaVA across various different tasks including classification, captioning and VQA to measure its robust generalization.

\subsection{Dual Adversarial Fine-tuning}
\label{sec:method detail}
To mitigate the overfitting phenomenon of adversarial fine-tuning and simultaneously improve robustness against multi-modal alignment attacks, we believe that all the generalizable features captured during pre-training, as well as the semantically rich supervision in adversarial fine-tuning, are especially valuable. 
The former helps maintain robustness in downstream tasks, while the latter is the key source of robustness. 
To achieve this, we propose Dual Adversarial Fine-tuning (DAFT), which utilizes dual supervision from both visual and semantic perspectives to achieve the objectives mentioned above.
In DAFT, the visual supervision branch utilizes the clean example features extracted by the frozen original vision encoder as supervision signals to preserve features beneficial for generalization and mitigate overfitting. 
In semantic supervision branch, caption-image alignment is used as a semantic supervision signal, fully exploring and leveraging the rich semantic information in the captions.
The pipeline of DAFT is shown in Fig. \ref{fig:pipeline}, and we will describe the design of DAFT in detail.

\noindent\textbf{Visual Supervision Branch.}
As previously analyzed, since adversarial examples are generated based on a specific dataset, the robustness features acquired by the target model are limited and may excessively overfit in a particular downstream task dataset, thus affecting the model's generalization ability. 
Visual supervision branch aims to alleviate this issue. 
The main body of this branch is an image encoder of the original CLIP, denoted as $F_{ori}(\cdot)$.
We denote $F_{ori}(x)$ and $F_{\theta}(x_a)$ as the embeddings of the clean example passing through the original model $F_{ori}$ and the adversarial example passing through the target model $F_{\theta}$, respectively.
Since the pre-trained model is a fixed deterministic function, the supervision objective of this branch essentially encourages the target model $F_{\theta}(\cdot)$ to output features that can predict the information of the original foundational model as much as possible, thereby helping to mitigate the overfitting problem.
This supervision branch is ultimately implemented using a loss based on $\mathcal{L}_2$ distance metric:
\begin{equation}
\begin{aligned}
\mathcal{L}_\text{vis} = ||F_{ori}(x)-F_{\theta}(x_a)||_2^2, 
\end{aligned}
\end{equation}
adversarial examples $x_a$ are dynamically generated during fine-tuning, with the update process detailed in Equ. \eqref{attack_daft}.

\noindent\textbf{Semantic Supervision Branch.}
The objective of semantic supervision is to ensure that adversarial examples $x_a$ remain semantically aligned with the text modality, thereby enhancing robustness against stronger LVLM adversarial attacks that aim to disrupt cross-modal alignment.
Two key issues should be addressed: (1) the choice of textual supervision, and (2) the strategy for alignment.
For issue 1, we introduce the use of descriptive captions as a novel form of text supervision. 
Unlike simple class labels (e.g., “a photo of [class]”), these captions offer richer semantic information, thereby enabling the generation of more effective adversarial examples. 
In addition, high-quality captions often include crucial contextual cues that are beneficial for tackling complex tasks such as Visual Question Answering (VQA).
For issue 2, we adopt a contrastive learning framework to align adversarial example $x_a$ with the ground-truth caption (i.e., positive sample), while increasing the distance between $x_a$ and other captions in the same batch (i.e., negative samples), thus achieving a more semantically rich alignment. 

Specifically, we use triplet loss to implement contrastive learning. 
We take the embedding of the adversarial example $x_a$, denoted as $F_{\theta}(x_a)$, as the anchor, and then use the frozen original CLIP text encoder $T_{\phi}(\cdot)$ to encode a batch of $N$ captions. 
One of these captions matches the clean image $x$ associated with the adversarial sample $x_a$, which serves as the positive sample, denoted as $Po$.
The selection of negative examples can vary, under the triplet loss implementation, we choose the ``hardest" sample from the batch as the negative example. 
The ``hardest" sample is defined as the caption among the remaining $N-1$ samples whose embedding is most similar to the embedding of the adversarial sample $F_{\theta}(x_a)$, which is denoted as $Ne$.
Finally, regarding the distance metric, we use cosine similarity, which is consistent with the original design of CLIP. The semantic supervision loss is implemented as follows:
\begin{equation}
\begin{split}
\mathcal{L}_\text{sem} = \max \Bigg(
    & \frac{F_{\theta}(x_a) \cdot T_{\phi}(Ne)}{\|F_{\theta}(x_a)\| \|T_{\phi}(Ne)\|} \\
    & - \frac{F_{\theta}(x_a) \cdot T_{\phi}(Po)}{\|F_{\theta}(x_a)\| \|T_{\phi}(Po)\|} 
    + a,\ 0
\Bigg)
\end{split}
\end{equation}
where $a$ is a hyper-parameter in the contrastive loss, used to control the learning difficulty and regulate the distance gap between the anchor and the positive/negative samples, and the generation of adversarial examples also follows the Equ. \eqref{attack_daft}.

\noindent\textbf{Adversarial Attack and Model Parameter Updates.}
Based on the supervision from the two modalities mentioned above, the final loss function of our method is:
\begin{equation}
\begin{aligned}
\label{totalloss}
\mathcal{L}_\text{DAFT} = \mathcal{L}_\text{sem} + \gamma \mathcal{L}_\text{vis}, 
\end{aligned}
\end{equation}
where $\gamma$ is the hyper-parameter balancing coefficient between two terms.
The process of adversarial fine-tuning involves a dynamic iteration between adversarial attacks and model parameter updates, with both alternating processes using the same objective function in Equ. \eqref{totalloss}, while in opposite optimization directions. 
Specifically, the optimization objective of the adversarial attack is: 
\begin{equation}
\begin{aligned}
\label{attack_daft}
x_a = \arg\max_{\lVert x_a - x \rVert_\infty \leq \varepsilon} \mathcal{L}_\text{DAFT}.
\end{aligned}
\end{equation}

We use the PGD algorithm \cite{madry2017towards}, as described in Equ. \eqref{pgd-cls}, to optimize Equ. \eqref{attack_daft}.
After obtaining the adversarial sample $x_a$, the model takes $x_a$ as input, and the update process is: 
\begin{equation}
\begin{aligned}
\theta = \arg\min_{\theta} \sum_{i=1}^{n} \mathcal{L}_{\text{DAFT}}(\theta, x_a^i),
\end{aligned}
\end{equation}
which is the specific implementation of Equ. \eqref{minmax game}.
$i$ represents the $i$-th sample, and $n$ represents the total number of training samples.

\section{Experiments}

\begin{table*}[ht]
\centering
\caption{Clean and Robust accuracy of the CLIP model on various image classification datasets (\%). The highest accuracy for each dataset under two attack budgets is highlighted in bold. The original CLIP model exhibits no robustness under adversarial attacks. On average robust accuracy across all datasets, the DAFT outperforms existing methods at both $\epsilon$ = 2/255 and $\epsilon$ = 4/255. In the last column, we report the training time required for each batch step of different methods (s/batch step).}
\label{tab:zs_eval}
\begin{adjustbox}{width=0.98\textwidth,keepaspectratio}
\begin{tabular}{c|lccccccccccc|c|c}
\hline
\multicolumn{1}{l|}{Perturbation Budget} & Method                               & Caltech101                   & Cars                         & CIFAR10                      & CIFAR100                     & DTD                          & EuroSAT                     & Fgvc                        & Flowers102                      & OxfordPet                         & STL10                          & ImageNet-r                   & Average   & Time                          \\ \hline
                                     & CLIP                                 & 82.7                         & 78.6                         & 95.7                         & 72.8                         & 55.7                         & 63.1                         & 32.7                         & 79.6                         & 92.5                         & 99.5                         & 87.4                         & 76.4           & -              \\
                                     & TeCOA \cite{mao2022understanding}                              & 70.3                         & 20.3                         & 60.0                           & 23.4                         & 18.2                         & 17.5                         & 7.2                          & 23.2                         & 63.3                         & 89.5                         & 51.1                         & 40.4          & 166               \\
                                     & PMG-AFT \cite{wang2024pre}                            & 70.8                         & 16.3                         & 59.6                         & 20.9                         & 24.3                         & 18.2                         & 7.3                          & 32.8                         & 58.5                         & 88.3                         & 43.9                         & 40.1              & 171           \\
                                     & FARE \cite{schlarmann2024robust}                               & 80.8                         & 61.9                         & 73.7                         & 41.2                         & 42.0                           & 10.1                         & 24.4                         & 55.8                         & 85.7                         & 93.4                         & 72.6                         & 58.3       & 159                  \\
\multirow{-5}{*}{clean}              & \cellcolor[HTML]{EFEFEF}DAFT (ours)       & \cellcolor[HTML]{EFEFEF}78.8 & \cellcolor[HTML]{EFEFEF}57.3 & \cellcolor[HTML]{EFEFEF}75.1 & \cellcolor[HTML]{EFEFEF}44.7 & \cellcolor[HTML]{EFEFEF}43.6 & \cellcolor[HTML]{EFEFEF}16.3 & \cellcolor[HTML]{EFEFEF}24.1 & \cellcolor[HTML]{EFEFEF}46.5 & \cellcolor[HTML]{EFEFEF}82.1 & \cellcolor[HTML]{EFEFEF}93.7 & \cellcolor[HTML]{EFEFEF}69.7 & \cellcolor[HTML]{EFEFEF}57.4  & 162 \\ \hline
                                     & CLIP                                 & 0.0                            & 0.0                            & 0.0                            & 0.0                            & 0.0                            & 0.0                           & 0.0                           & 0.0                            & 0.0                            & 0.0                            & 0.0                            & 0.0                & -                  \\
                                     & TeCOA \cite{mao2022understanding}                              & 58.8                         & 7.9                          & 38.0                           & 12.4                         & 11.3                         & 11.1                        & 2.7                         & 9.2                          & 38.3                         & 73.3                         & 32.9                         & 26.9             & 166                  \\
                                     & PMG-AFT \cite{wang2024pre}                           & 57.3                         & 6.1                          & 37.9                         & 11.3                         & 15.1                         & \textbf{11.3}                        & 3.4                         & 16.6                         & 33.7                         & 72.3                         & 26.3                         & 26.5               & 171            \\
                                     & FARE \cite{schlarmann2024robust}                       & \textbf{70.1}                         & \textbf{23.5}                         & 49.9                         & 23.4                         & 22.3                         & 3.7                         & 8.4                         & \textbf{27.1}                         & 60.3                         & \textbf{79.7}                         & 47.0                           & 37.7                     & 159      \\
\multirow{-5}{*}{$\epsilon$=2/255}            & \cellcolor[HTML]{EFEFEF}DAFT (ours) & \cellcolor[HTML]{EFEFEF}69.8 & \cellcolor[HTML]{EFEFEF}22.6 & \cellcolor[HTML]{EFEFEF}\textbf{53.4} & \cellcolor[HTML]{EFEFEF}\textbf{25.8} & \cellcolor[HTML]{EFEFEF}\textbf{24.7} & \cellcolor[HTML]{EFEFEF}9.9 & \cellcolor[HTML]{EFEFEF}\textbf{9.1} & \cellcolor[HTML]{EFEFEF}24.6 & \cellcolor[HTML]{EFEFEF}\textbf{61.7} & \cellcolor[HTML]{EFEFEF}79.6 & \cellcolor[HTML]{EFEFEF}\textbf{47.2} & \cellcolor[HTML]{EFEFEF}\textbf{38.9} & 162  \\ \hline
                                     & CLIP                                 & 0.0                            & 0.0                            & 0.0                            & 0.0                            & 0.0                            & 0.0                           & 0.0                           & 0.0                            & 0.0                            & 0.0                            & 0.0                            & 0.0                     & -             \\
                                     & TeCOA \cite{mao2022understanding}                              & 46.2                         & 3.5                          & 21.0                           & 6.7                          & 6.4                          & 10.5                        & 0.8                         & 3.0                            & 14.8                         & 50.6                         & 18.0                           & 16.5       & 166                      \\
                                     & PMG-AFT \cite{wang2024pre}                            & 43.9                         & 2.5                          & 21.7                         & 6.2                          & 10.6                         & \textbf{11.4}                        & 1.0                           & 7.4                          & 14.6                         & 49.6                         & 16.1                         & 16.8                           & 171\\
                                     & FARE \cite{schlarmann2024robust}                               & 56.0                           & 4.5                          & 26.7                         & 12.8                         & \textbf{13.9}                         & 6.8                         & 2.4                         & 4.3                          & 26.4                         & 55.9                         & 27.9                         & 21.6                               & 159\\
\multirow{-5}{*}{$\epsilon$=4/255}            & \cellcolor[HTML]{EFEFEF}DAFT (ours)       & \cellcolor[HTML]{EFEFEF}\textbf{56.4} & \cellcolor[HTML]{EFEFEF}\textbf{7.1}  & \cellcolor[HTML]{EFEFEF}\textbf{28.1} & \cellcolor[HTML]{EFEFEF}\textbf{13.2} & \cellcolor[HTML]{EFEFEF}12.4 & \cellcolor[HTML]{EFEFEF}4.4 & \cellcolor[HTML]{EFEFEF}\textbf{2.8} & \cellcolor[HTML]{EFEFEF}\textbf{7.9}  & \cellcolor[HTML]{EFEFEF}\textbf{32.0}   & \cellcolor[HTML]{EFEFEF}\textbf{59.8} & \cellcolor[HTML]{EFEFEF}\textbf{29.4} & \cellcolor[HTML]{EFEFEF}\textbf{23.0} & 162  \\ \hline
\end{tabular}
\end{adjustbox}
\end{table*}

\subsection{Experimental Setup}
\label{setup}
\noindent\textbf{Models and Tasks.}
We adopt the LLaVA \cite{liu2024improved} framework to explore robustness of large vision-language models, which uses CLIP ViT-L-14 \cite{radford2021learning} as its vision encoder.
In zero-shot classification tasks, inference can be performed using only the alignment mechanism of CLIP itself as introduced in Sec. \ref{Preliminaries}. 
However, for tasks like image captioning and VQA, we replace the original encoder in LLaVA with a trained robust vision encoder for downstream task inference.

\noindent\textbf{Datasets and Metrics.} 
We fine-tune the CLIP model on the COCO-caption \cite{lin2014microsoft} dataset and evaluate it on a wide range of downstream tasks across various datasets. 
For zero-shot classification tasks, we select 11 datasets: Caltech101 \cite{griffin2007caltech}, StanfordCars \cite{krause20133d}, CIFAR10 \cite{krizhevsky2009learning}, CIFAR100 \cite{krizhevsky2009learning}, DTD \cite{cimpoi2014describing}, EuroSAT \cite{helber2019eurosat},  FGVC Aircrafts \cite{maji2013fine}, Flowers \cite{nilsback2008automated}, OxfordPets \cite{parkhi2012cats}, STL-10 \cite{coates2011analysis}, and ImageNet-R \cite{hendrycks2021many}. 
For image captioning, we use the COCO \cite{lin2014microsoft} and Flickr30k \cite{plummer2015flickr30k} datasets. 
For visual question answering, we use the VQAv2 \cite{goyal2017making}, TextVQA \cite{singh2019towards}, VizWiz \cite{gurari2018vizwiz}, and OKVQA \cite{marino2019ok} datasets.
For zero-shot classification and VQA tasks, we evaluate the adversarial sample classification accuracy and answer accuracy, respectively. 
For the image captioning task, we report the CIDEr scores \cite{vedantam2015cider} between the responses to adversarial examples and the ground-truth captions.

Due to computational resource limitations, we use a relatively smaller COCO-caption \cite{lin2014microsoft} dataset for training. 
Although the scale of COCO \cite{lin2014microsoft} is smaller than ImageNet \cite{deng2009imagenet} or the larger LAION-400M \cite{schuhmann2022laion} dataset, it has unique advantages. 
First, COCO images are more complex than those in ImageNet and contain richer textual information, which aids in image-text contrastive learning and enables the model to learn more robust features. 
Additionally, ImageNet currently lacks mature caption annotations, which further highlights COCO's value. 
Second, the quality of annotations in the COCO dataset is relatively high and accurately reflects the content of the images, making it a more reliable choice compared to the broad, crowdsourced image-text matching datasets like LAION.

To ensure fair comparison in our experiments, all competing methods (FARE \cite{schlarmann2024robust}, TeCOA \cite{mao2022understanding}) are trained on the same COCO dataset. 
This is feasible since, apart from the caption annotations, COCO also contains category labels that can be adapted for other methods. 
As a result, the training and testing datasets are consistent across models, ensuring fairness throughout the experiments.

\noindent\textbf{Baseline.} 
As mentioned in the related work, adversarial robustness in the LVLM field is still in the early stages of research. 
We mainly compare our method with category-label supervised methods such as TeCOA \cite{mao2022understanding}, PMG-AFT \cite{wang2024pre}, and another unsupervised method, FARE \cite{schlarmann2024robust}, which is currently the state-of-the-art (SOTA) method. 
We also maintain the same settings as FARE \cite{schlarmann2024robust} for the experiments.

\noindent\textbf{Implementation Details.} 
Since our method uses caption data as text supervision, we select the popular COCO dataset as the training set, which also provides category labels, allowing it to support the training of methods like TeCoA \cite{mao2022understanding}.
Noted that, for fair comparison, all other methods compared in this paper are also trained on the COCO dataset.
To remain consistent with LLaVA's backbone, we adopt the CLIP model's ViT-L/14 architecture as our target model.
We use the AdamW \cite{loshchilov2017decoupled} optimizer, with momentum coefficients $\beta_1$ and $\beta_2$ set to 0.9 and 0.95, respectively. 
The training utilizes a cosine decay learning rate (LR) schedule, with a linear warmup to the peak LR 1e-5. 
Weight decay is set to 1e-4, and the  batch size is 64.
During training, we use PGD-10 \cite{madry2017towards} attacks with $\mathcal{L}_\infty$ norm perturbation budget of $\epsilon$ = 4/255. 
For hyper-parameters, we set $\gamma$ to 0.5 and $a$ to 0.2.
Each model in this paper is trained on two NVIDIA A100 GPUs.
During evaluation, we use LLaVA1.5-7B \cite{liu2024improved} as the LVLM for our evaluation, and we apply different attack algorithms depending on the task, as detailed in Sec. \ref{zero-shot} and \ref{vqa-cap}.

\subsection{Evaluation of Zero-Shot Classification}
\label{zero-shot}
\noindent\textbf{Attack Implementation.} 
We evaluate robust accuracy on 1000 randomly sampled instances from each dataset mentioned in Sec. \ref{setup}. 
We use AutoAttack \cite{croce2020reliable}, consisting of APGD with cross-entropy loss and APGD with targeted DLR loss, to perform 100 iterations attacks.
Models are tested under $\mathcal{L}_\infty$-bounded adversarial perturbations with budgets of $\epsilon$ = 2/255 and $\epsilon$ = 4/255.
Since all models are trained exclusively on the COCO dataset, classification performance across various downstream datasets is evaluated in a zero-shot manner.

\noindent\textbf{Main Results.} 
As shown in Tab. \ref{tab:zs_eval}, the original CLIP model without adversarial fine-tuning achieves superior clean accuracy but unsurprisingly exhibits no adversarial robustness.
In contrast, Our DAFT outperforms existing methods in terms of robust accuracy across most datasets.
Compared to category-label supervised methods, e.g., TeCoA \cite{mao2022understanding} and PMG-AFT \cite{wang2024pre}, DAFT improves average robust accuracy by  approximately 12\% and 6.5\% under perturbation budgets of 2/255 and 4/255, respectively, while maintaining a significantly smaller degradation in clean accuracy. 
This aligns with our claim in Sec. \ref{intro} that such methods are \textbf{prone to overfitting and lack generalization}. 
In comparison to the unsupervised method FARE \cite{schlarmann2024robust}, our method improves average robustness gains of 1.2\% and 1.4\% under perturbation budgets of 2/255 and 4/255, respectively, suggesting that while FARE partially enhances generalization, it does not fully optimize robust performance.
Notably, DAFT’s clean accuracy remains comparable to FARE, albeit marginally lower, a trade-off we attribute to the introduction of caption supervision. 
This may be due to the introduction of additional caption supervision, which increases the deviation between adversarial samples and clean samples, leading to a slight decline in clean accuracy.
Overall, DAFT demonstrates superior robust generalization for zero-shot classification tasks, balancing adversarial robustness and clean accuracy more effectively than existing paradigms.

\subsection{Performance on Image Captioning and VQA}
\label{vqa-cap}

\begin{table*}[ht]
\centering
\caption{Clean and Robust Performance of LLaVA with different vision encoders on different captioning and VQA datasets.
For caption datasets Flickr30 and COCO, we report their CIDEr scores. For other four VQA datasets, we report their VQA accuracy.
The highest accuracy for each dataset under two attack budgets is highlighted in bold.
Overall, DAFT demonstrates the best robust performance compared to other methods while achieving comparable clean performance.}
\label{tab:capvqa}
\begin{adjustbox}{width=0.98\textwidth,keepaspectratio}
\begin{tabular}{c|l|cccccc|c|c}
\hline
\multicolumn{1}{l|}{Perturbation Budget} & Model                                     & Flickr30                              & COCO                                  & VQAv2                                 & OKVQA                                 & VizWiz                                & TextVQA                               & Average-Caption                           & Average-VQA                           \\ \hline
                                         & LLaVA-CLIP                                & 83.2                                  & 121.6                                 & 76.7                                  & 60.6                                  & 38.8                                  & 45.3                                  & 102.4                                 & 55.3                                  \\
                                         & LLaVA-TeCOA \cite{mao2022understanding}                              & 38.6                                  & 90.9                                  & 61.4                                  & 51.2                                  & 42.5                                  & 17.0                                  & 64.7                                  & 43.0                                  \\
                                         & LLaVA-PMG-AFT \cite{wang2024pre}                            & 40.4                                  & 91.8                                  & 60.3                                  & 50.7                                  & 42.4                                  & 16.7                                  & 66.1                                  & 42.5                                  \\
                                         & LLaVA-FARE  \cite{schlarmann2024robust}                              & 57.7                                  & 107.8                                 & 67.6                                  & 55.9                                  & 42.6                                  & 26.4                                  & 82.7                                  & 48.1                                  \\
\multirow{-5}{*}{clean}                  & \cellcolor[HTML]{EFEFEF}LLaVA-DAFT (ours) & \cellcolor[HTML]{EFEFEF}56.3          & \cellcolor[HTML]{EFEFEF}105.3         & \cellcolor[HTML]{EFEFEF}66.7          & \cellcolor[HTML]{EFEFEF}55.8          & \cellcolor[HTML]{EFEFEF}42.4          & \cellcolor[HTML]{EFEFEF}25.2          & \cellcolor[HTML]{EFEFEF}80.8          & \cellcolor[HTML]{EFEFEF}47.5          \\ \hline
                                         & LLaVA-CLIP                                & 3.9                                   & 7.9                                   & 7.7                                   & 1.6                                   & 0.1                                   & 0.8                                   & 5.9                                   & 2.5                                   \\
                                         & LLaVA-TeCOA \cite{mao2022understanding}                              & 12.0                                  & 41.3                                  & 38.3                                  & 28.5                                  & 29.0                                  & 8.5                                   & 26.6                                  & 26.1                                  \\
                                         & LLaVA-PMG-AFT  \cite{wang2024pre}                           & 11.9                                  & 40.6                                  & 36.6                                  & 29.1                                  & 29.2                                  & 8.7                                   & 26.3                                  & 25.9                                  \\
                                         & LLaVA-FARE  \cite{schlarmann2024robust}                              & 21.4                                  & 56.9                                  & 40.2                                  & 32.8                                  & 30.2                                  & \textbf{12.6}                         & 39.2                                  & 29.0                                  \\
\multirow{-5}{*}{$\epsilon$=2/255}                & \cellcolor[HTML]{EFEFEF}LLaVA-DAFT (ours) & \cellcolor[HTML]{EFEFEF}\textbf{23.0} & \cellcolor[HTML]{EFEFEF}\textbf{59.9} & \cellcolor[HTML]{EFEFEF}\textbf{41.4} & \cellcolor[HTML]{EFEFEF}\textbf{34.9} & \cellcolor[HTML]{EFEFEF}\textbf{31.4} & \cellcolor[HTML]{EFEFEF}\textbf{12.6} & \cellcolor[HTML]{EFEFEF}\textbf{41.5} & \cellcolor[HTML]{EFEFEF}\textbf{30.1} \\ \hline
                                         & LLaVA-CLIP                                & 2.1                                   & 4.1                                   & 1.5                                   & 0.1                                   & 0.1                                   & 0.0                                   & 3.1                                   & 0.4                                   \\
                                         & LLaVA-TeCOA  \cite{mao2022understanding}                             & 8.8                                   & 32.9                                  & 30.7                                  & 23.4                                  & 25.0                                  & 6.2                                   & 20.9                                  & 21.3                                  \\
                                         & LLaVA-PMG-AFT  \cite{wang2024pre}                           & 8.9                                   & 31.1                                  & 29.7                                  & 23.1                                  & 24.8                                  & 5.9                                   & 20.0                                  & 20.9                                  \\
                                         & LLaVA-FARE  \cite{schlarmann2024robust}                              & 16.1                                  & 45.2                                  & 33.0                                  & 26.0                                  & 25.9                                  & \textbf{9.4}                          & 30.6                                  & 23.6                                  \\
\multirow{-5}{*}{$\epsilon$=4/255}                & \cellcolor[HTML]{EFEFEF}LLaVA-DAFT (ours) & \cellcolor[HTML]{EFEFEF}\textbf{17.3} & \cellcolor[HTML]{EFEFEF}\textbf{48.1} & \cellcolor[HTML]{EFEFEF}\textbf{34.5} & \cellcolor[HTML]{EFEFEF}\textbf{27.9} & \cellcolor[HTML]{EFEFEF}\textbf{27.0} & \cellcolor[HTML]{EFEFEF}8.7           & \cellcolor[HTML]{EFEFEF}\textbf{32.7} & \cellcolor[HTML]{EFEFEF}\textbf{24.5} \\ \hline
\end{tabular}
\end{adjustbox}
\end{table*}
\noindent\textbf{Attack Implementation.} 
We evaluate the performance of LLaVA 1.5-7B under untargeted attack scenarios when using the original CLIP vision encoder versus robust CLIP vision encoders trained through different adversarial fine-tuning methods.
Consistent with previous work \cite{schlarmann2023adversarial,schlarmann2024robust}, we adopt a multi-stage attack ensemble strategy. 
For each dataset, we randomly select 1000 samples and perform $\mathcal{L}_\infty$-bounded attacks with two different perturbation budgets of $\epsilon$ = 2/255 and $\epsilon$ = 4/255. 
In the first stage, we apply half-precision (FP16) APGD attacks \cite{croce2020reliable} for 100 iterations to all selected samples. 
We then retain only those samples for which the CIDEr score exceeds a certain threshold for image captioning tasks, or those that were not successfully attacked for VQA tasks, subjecting them to a second stage of single-precision (FP32) APGD attacks for 100 iterations. 
This attack strategy optimizes computational resources while enhancing attack effectiveness, as verified in \cite{schlarmann2024robust}.

\noindent\textbf{Main Results.} 
Tab. \ref{tab:capvqa} presents the evaluation results of adversarial robustness across 6 datasets. 
LLaVA using the original CLIP achieves the best clean performance but is completely vulnerable to adversarial attacks. 
Among adversarially fine-tuning models, FARE \cite{schlarmann2024robust} demonstrates the best clean performance, similar to its performance on zero-shot classification task. 
This improvement may come from explicitly aligning the robust CLIP model's representations with the original CLIP, which limits the diversity of adversarial attacks and allows the adversarially fine-tuned model to predict clean sample outputs to the greatest extent.
This also leads to FARE not achieving the best results under attacks targeting vision-language tasks, indirectly supporting the second challenge we raised in Sec. \ref{intro}: \textbf{Robustness Gains are Limited from Unsupervised Adversarial Fine-Tuning.}
Moreover, category label-supervised methods like TeCoA \cite{mao2022understanding} and PMG-AFT \cite{wang2024pre} underperform in both clean and adversarial scenarios. 
In contrast, DAFT significantly improves adversarial robustness with minimal clean-performance degradation, demonstrating an effective trade-off between clean and adversarial generalization.
Specifically, under a perturbation budget of $\epsilon$ = 2/255, DAFT outperforms TeCoA in robustness across all 6 datasets, with an average increase of 14.9 in CIDEr scores and 4.0\% improvement in average VQA accuracy. 
Compared to FARE, DAFT improves the average CIDEr score by 2.3 and the average VQA accuracy by 1.1\%.
Under stronger perturbations of $\epsilon$ = 4/255, DAFT improves the average caption CIDEr score by 11.8 and the average VQA accuracy by 3.2\% compared to TeCoA. 
Compared to FARE, DAFT improves the average CIDEr score by 2.1 and the average VQA accuracy by 0.9\%.

\subsection{Ablation Study}

\noindent\textbf{Contribution of Loss Function Term.} 
To demonstrate the effectiveness of the dual-supervision loss, we incrementally integrate individual loss components during training, evaluating them under attacks with a perturbation budget of $\epsilon$ = 2/255. 
The attack methods remain consistent with those in Sec. \ref{zero-shot} and \ref{vqa-cap}, and we report their average robust performance across three tasks.
Additionally, to highlight the superiority of captions over category label as supervision, we conduct an extra experiment where we replace the $\mathcal{L}_\text{sem}$ with category label supervision as TeCOA \cite{mao2022understanding} (shown in the first row of the Tab. \ref{tab:lossterm}). 
As shown in Tab. \ref{tab:lossterm}, models trained solely with visual supervision or semantic supervision exhibit suboptimal robustness. 
Among them, the performance of using semantic supervision alone decreases more significantly, suggesting that while captions offer some generalization ability, their isolated use still suffers from severe overfitting. 
On the other hand, using visual supervision alone similarly fails to maximize robustness, indicating that the two losses complement and promote each other, further underscoring the necessity of our dual adversarial fine-tuning framework.
When replacing captions with category labels as supervision, the average zero-shot classification accuracy decreases by 12.4\%, the average caption CIDEr score drops by 15.2, and the average VQA accuracy decreases by 4.2\% (as illustrated in the first and last rows of Tab. \ref{tab:lossterm}), highlighting the superiority of caption-guided semantic alignment over class-label constraints, validating our design rationale for DAFT.

\begin{table}[htbp]
\centering
\caption{Contribution of each term in the loss function. 
We report robustness performance under a perturbation budget of $\epsilon$=2/255 across three tasks. 
Noted that the first row in the table represents using category label supervision as TeCOA \cite{mao2022understanding} to replace the semantic supervision $\mathcal{L}_\text{sem}$ we propose.}
\label{tab:lossterm}
\begin{adjustbox}{width=0.48\textwidth,keepaspectratio}
\begin{tabular}{lll|ccc}
\hline
$\mathcal{L}_\text{sem}$          & $\mathcal{L}_\text{vis}$  &        & Avg-zeroshot-cls \ (\%) & Avg-caption \ (CIDEr) & Avg-VQA \ (\%) \\ \hline
category-sup  & \checkmark & & 26.5            & 26.3            & 25.9        \\ 

\checkmark           &  &          & 33.7            & 30.1            & 24.2          \\
            & \checkmark &         & 37.7            & 39.2            & 29.0        \\
\checkmark           & \checkmark & (DAFT)         & \textbf{38.9}             & \textbf{41.5}            & \textbf{30.1}       \\ \hline
\end{tabular}
\end{adjustbox}
\end{table}

\begin{table}[htbp]
\centering
\caption{The impact of different implementations of contrastive loss and sampling strategies in semantic supervision branch.
We report robustness performance under a perturbation budget of $\epsilon$=2/255 across three tasks. }
\label{tab:sampling}
\begin{adjustbox}{width=0.48\textwidth,keepaspectratio}
\begin{tabular}{l|ccc}
\hline
Sampling Strategy              & Avg-zeroshot-cls  \ (\%) & Avg-caption \ (CIDEr) & Avg-VQA \ (\%) \\ \hline
Triplet loss  (DAFT)           & \textbf{38.9}                                 & \textbf{41.5}                               & \textbf{30.1}                        \\
Triplet loss  (easiest sample) & 0.2                                  & 6.7                                & 3.4                         \\
InfoNCE loss                   & 35.2                                 & 41.2                               & 29.8                        \\
Quadruplet loss (second)        & 38.0                                   & 36.3                               & 29.5                        \\
Quadruplet loss (random)       & 38.4                                 & 34.0                                 & 28.5                        \\ \hline
\end{tabular}
\end{adjustbox}
\end{table}

\noindent\textbf{Impact of Negative Sampling Strategy.} 
In contrastive learning, the positive and negative sample sampling strategy is a crucial factor that influences model performance. 
We implement the loss $\mathcal{L}_\text{sem}$ using different sampling strategies for contrastive learning in semantic supervision branch.
Since positive samples are inherently defined by ground-truth image-caption pairs, we focus on how negative sample selection impacts adversarial robustness.
As shown in Tab. \ref{tab:sampling}, we compare two types of triplet loss. 
The first type, as in DAFT, selects the caption from the $N-1$ captions (excluding the positive sample) that is most similar to the adversarial features as the negative example. 
The second type selects the caption that is least similar to the adversarial features as the negative example (as shown in Tab. \ref{tab:sampling} ``easiest sample"). 
Strikingly, models trained with ``easiest negative" sampling exhibit near-zero robustness, mirroring the original CLIP’s vulnerability. 
This aligns with expectation: optimizing toward minimally conflicting negatives prematurely saturates the contrastive objective, halting meaningful model adaptation.
Additionally, we perform the InfoNCE loss to implement $\mathcal{L}_\text{sem}$. 
The key difference compared to triplet loss is that InfoNCE treats all $N-1$ captions as negative examples and maximizes the distance between the adversarial sample and these captions. 
The model trained with InfoNCE loss is almost comparable to DAFT in zero-shot classification tasks but shows some decrease in robustness in image captioning and VQA tasks.

To more thoroughly explore the adversarial example space, we extend the triplet loss to a quadruplet loss that incorporates two distinct negative examples. 
One negative example is fixed and consistent with DAFT: the caption that is most similar to the adversarial features from the $N-1$ captions. 
The second negative example is implemented in two different ways: one selects randomly from the remaining $N-2$ captions, and the other selects the caption that is most similar to the adversarial features from the remaining $N-2$ captions (``second" in Tab. \ref{tab:sampling}). 
Experimental results show that the model fine-tuned with quadruplet loss slightly reduces robustness in the image captioning, while performing similarly to DAFT in other tasks. 
This suggests that, although increasing the number of negative examples intuitively helps to fully explore the adversarial space and generate stronger attacks, the need to simultaneously increase the distance from these negative examples causes conflicts in the optimization direction. 
As a result, the model struggles to find the optimal solution, leading to a decrease in robustness.
Overall, the negative sample sampling strategy that includes the caption most similar to the adversarial feature proves to be the optimal choice.
As we have analyzed, such negative sample can be considered the ``hardest" sample, which helps improve robustness.

\begin{table}[t]
\centering
\caption{Average robustness performance across different tasks under a perturbation budget of $\epsilon$=2/255 for various hyper-parameter combinations.}
\label{tab:hyper}
\begin{adjustbox}{width=0.48\textwidth,keepaspectratio}
\begin{tabular}{l|ccc}
\hline
Hyper-Parameters & Avg-zeroshot-cls \ (\%) & Avg-caption \ (CIDEr) & Avg-VQA \ (\%) \\ \hline
$a$=0.1 \ \ $\gamma$=0.5 & 38.8                                 & 40.1                               & 29.9                        \\
$a$=0.2 \ \ $\gamma$=0.5 & \textbf{38.9}                        & \textbf{41.5}                      & \textbf{30.1}               \\
$a$=0.3 \ \ $\gamma$=0.5 & 38.4                                 & 41.2                               & 29.2                        \\
$a$=0.5 \ \ $\gamma$=0.5 & 38.1                                 & 41.1                               & 30.1                        \\ \hline
\rowcolor[HTML]{EFEFEF} 
$a$=0.2 \ \ $\gamma$=0.1 & 36.4                                 & 38.3                               & 29.5                        \\
\rowcolor[HTML]{EFEFEF} 
$a$=0.2 \ \ $\gamma$=0.2 & 37.1                                 & 39.9                               & 29.8                        \\
\rowcolor[HTML]{EFEFEF} 
$a$=0.2 \ \ $\gamma$=0.5 & \textbf{38.9}                        & \textbf{41.5}                      & \textbf{30.1}               \\
\rowcolor[HTML]{EFEFEF} 
$a$=0.2 \ \ $\gamma$=1   & 38.5                                 & 40.4                               & 29.6                        \\
\rowcolor[HTML]{EFEFEF} 
$a$=0.2 \ \ $\gamma$=5   & 38.0                                   & 39.4                               & 29.4                        \\ \hline
\end{tabular}
\end{adjustbox}
\end{table}

\noindent\textbf{Effect of hyper-parameters $\gamma$ and $a$.} 
Our dual-supervision loss comprises two terms $\mathcal{L}_\text{sem}$ and $\mathcal{L}_\text{vis}$, balanced by a hyper-parameter $\gamma$. 
Additionally, in the semantic loss $\mathcal{L}_\text{sem}$, a hyper-parameter $a$ denotes the margin in the triplet loss. 
Since we use the cosine similarity, the range of $a$ is between 0 and 1. 
To examine the impact of these two parameters on the model's robustness, we train models by gradually varying each of these hyper-parameters. 
As shown in Tab. \ref{tab:hyper}, when $\gamma$ is fixed at 0.5, as $a$ gradually increases, the model's overall robust performance first improves and then decreases. 
Since $a$ mainly adjusts the difficulty of contrastive learning, as $a$ increases, initially, increasing the difficulty of learning helps better align adversarial examples with the correct captions while pushing away negative captions. 
However, when $a$ becomes too large, it makes the model difficult to optimize, leading it to fall into a local minimum. 
We find that the model's robustness performance is the best when $a$=0.2.
When $a$ is fixed at 0.2, as $\gamma$ gradually increases, the model's overall robust performance also shows a trend of first improving and then decreasing. 
The model's robustness is best when $\gamma$=0.5, which indicates that two terms in the loss function are of similar importance, further demonstrating the effectiveness of DAFT.

\begin{figure}[t]
    \centering
    \includegraphics[scale=0.32]{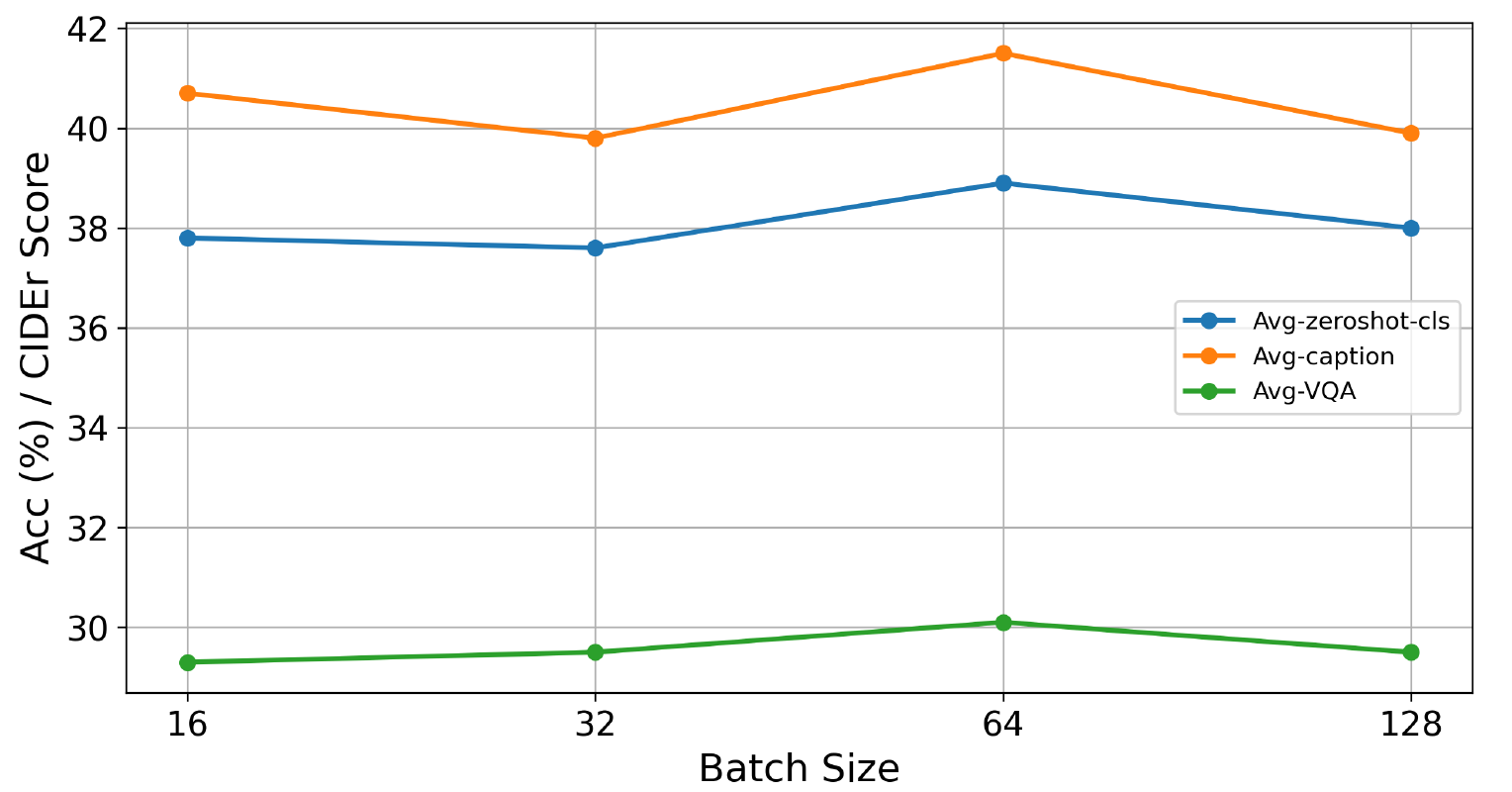}
    \caption{Average robustness performance of models trained with DAFT under different batch sizes across various tasks.}
    \label{fig:batchsize}
\end{figure}

\noindent\textbf{Effect of Batch Size.}
The effectiveness of $\mathcal{L}_\text{sem}$ hinges on contrastive learning dynamics, where batch size governs the diversity and difficulty of negative samples. 
We conduct a systematic evaluation of DAFT across different batch sizes under a perturbation budget of $\epsilon$=2/255, as shown in Fig. \ref{fig:batchsize}. 
The left vertical axis in the figure represents different meanings for different tasks: for zero-shot classification and VQA tasks, it represents the robust accuracy, while for the image captioning task, it denotes the CIDEr score. 
From the Fig. \ref{fig:batchsize}, we observe that as the batch size increases (from 16 to 128), the model's robustness performance across tasks consistently improves and then decreases. 
The best robustness performance is achieved when the batch size is 64.
The batch size primarily affects the selection of negative examples $Ne$. 
When the batch size is large, the likelihood of encountering captions in the same batch with high similarity to the adversarial features increases, making the selected negative examples more similar to the adversarial features. 
Once this similarity is close to that between the adversarial and the true caption, it makes contrastive learning difficult to optimize. 
On the other hand, if the batch size is small, the selected negative examples might be too simple. 
As shown in Tab. \ref{tab:sampling}, overly simple negative examples prevent contrastive learning from optimizing. 
Therefore, the batch size should neither be too large nor too small. 
We choose a batch size of 64 as the final implementation.

\section{Limitations and Future Work}
This work primarily focuses on CLIP-based large vision-language models (LVLMs). Other LVLMs that do not rely on CLIP as the vision encoder may also benefit from our robustness-enhancing approach, but we leave the empirical validation of this to future work. Secondly, our method is a white-box defense approach, which is not directly applicable to currently closed-source models. In addition, our method only fine-tunes the visual encoder, and the defense of the language component in LVLMs is also left for future exploration. Finally, since COCO as a caption dataset is limited in scale, we plan to further improve the quality of caption data in future work to enhance the performance of our method.


\section{Conclusion}
In this work, we analyze two limitations in current adversarial defense research for large vision-language models (LVLMs): overfitting and insufficient alignment with text. 
To address these issues, we propose a dual-supervision adversarial fine-tuning framework, which introduces descriptive captions for semantic supervision, enabling the model to better handle advanced vision-language attacks, while  preserving generalization through visual supervision.
Our framework fine-tunes the vision encoder of the CLIP model, allowing it to replace the original vision encoder in LVLM and thus achieve robustness across various multimodal tasks. 
Extensive experiments demonstrate that our method exhibits strong adversarial robustness across different datasets for downstream tasks, while maintaining high clean performance.

{
    \small
    \bibliographystyle{IEEEtran}
    \bibliography{main}
}

 \begin{IEEEbiography}[{\includegraphics[width=1in,height=1.25in,clip,keepaspectratio]{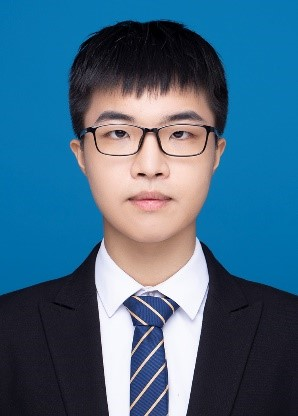}}]{Sibo Wang}
(Student Member, IEEE) received the B.S. degree from Harbin Institute of Technology in 2022. He is currently pursuing the Ph.D. degree from University of Chinese Academy of Sciences. His research interest includes adversarial example and model robustness. He has authored several academic papers in international conferences including CVPR/NIPS. 
\end{IEEEbiography}

\begin{IEEEbiography}[{\includegraphics[width=1in,height=1.25in,clip,keepaspectratio]{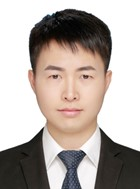}}]{Jie Zhang}
(Member, IEEE) received the Ph.D. degree from the University of Chinese Academy of Sciences (CAS), Beijing, China. He is currently an Associate Professor with the Institute of Computing Technology, CAS. His research interests include computer vision, pattern recognition, machine learning, particularly include face recognition, image segmentation, weakly/semi-supervised learning, and domain generalization.
\end{IEEEbiography}

\begin{IEEEbiography}[{\includegraphics[width=1in,height=1.25in,clip,keepaspectratio]{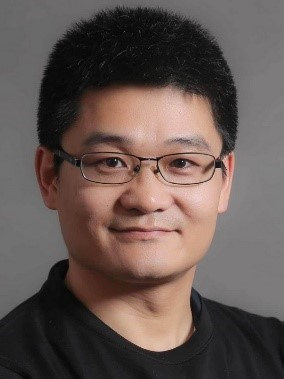}}]{Shiguang Shan}
(Fellow, IEEE) received the Ph.D. degree in computer science from the Institute of Computing Technology (ICT), Chinese Academy of Sciences (CAS), Beijing, China, in 2004. He has been a Full Professor with ICT since 2010, where he is currently the Director of the Key Laboratory of Intelligent Information Processing, CAS. His research interests include signal processing, computer vision, pattern recognition, and machine learning. He has published more than 300 articles in related areas. He served as the General Co-Chair for IEEE Face and Gesture Recognition 2023, the General Co-Chair for Asian Conference on Computer Vision (ACCV) 2022, and the Area Chair of many international conferences, including CVPR, ICCV, AAAI, IJCAI, ACCV, ICPR, and FG. He was/is an Associate Editors of several journals, including IEEE Transactions on Image Processing, Neurocomputing, CVIU, and PRL. He was a recipient of the China's State Natural Science Award in 2015 and the China’s State S\&T Progress Award in 2005 for his research work.
\end{IEEEbiography}

\begin{IEEEbiography}[{\includegraphics[width=1in,height=1.25in,clip,keepaspectratio]{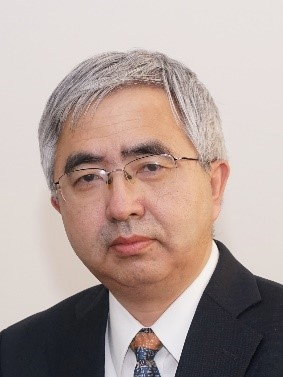}}]{Xilin Chen}
(Fellow, IEEE) is currently a Professor with the Institute of Computing Technology, Chinese Academy of Sciences (CAS). He has authored one book and more than 400 articles in refereed journals and proceedings in the areas of computer vision, pattern recognition, image processing, and multimodal interfaces. He is a fellow of the ACM, IAPR, and CCF. He is also an Information Sciences Editorial Board Member of Fundamental Research, an Editorial Board Member of Research, a Senior Editor of the Journal of Visual Communication and Image Representation, and an Associate Editor-in-Chief of the Chinese Journal of Computers and Chinese Journal of Pattern Recognition and Artificial Intelligence. He served as an organizing committee member for multiple conferences, including the General Co-Chair of FG 2013/FG 2018, VCIP 2022, the Program Co-Chair of ICMI 2010/FG 2024, and an Area Chair of ICCV/CVPR/ECCV/NeurIPS for more than ten times.
\end{IEEEbiography}

\begin{IEEEbiography}[{\includegraphics[width=1in,height=1.25in,clip,keepaspectratio]{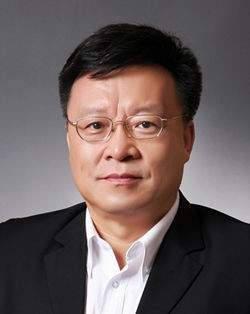}}]{Wen Gao}
(Fellow, IEEE) received the Ph.D. degree in electronics engineering from The University of Tokyo, Japan, in 1991. He is currently a Boya Chair Professor in computer science at Peking University. He is also the Director of the Peng Cheng Laboratory, Shenzhen. Before joining Peking University, he was a Professor with the Harbin Institute of Technology, from 1991 to 1995. From 1996 to 2006, he was a Professor with the Institute of Computing Technology, Chinese Academy of Sciences. He has published extensively, including five books and over 1000 technical articles in refereed journals and conference proceedings in the areas of image processing, video coding and communication, computer vision, multimedia retrieval, multimodal interface, and bioinformatics. He served on the editorial boards for several journals, such as ACM CSUR, IEEE Transactions on Image Processing (TIP), IEEE Transactions on Circuits and Systems for Video Technology (TCSVT), and IEEE Transactions on Multimedia (TMM). He served on the advisory and technical committees for professional organizations. He was the Vice President of the National Natural Science Foundation (NSFC) of China, from 2013 to 2018, and the President of China Computer Federation (CCF), from 2016 to 2020. He is also the Deputy Director of China National Standardization Technical Committees. He is an Academician of the Chinese Academy of Engineering and a fellow of ACM.
\end{IEEEbiography}

\clearpage
\section*{Supplementary Material}
\setcounter{section}{0}

\section{Additional Results under Stronger Attacks}

To further evaluate the robustness of different vision encoders under more challenging adversarial perturbations, we provide additional results with a larger perturbation budget of $\epsilon$ = 8/255 in Tab.~\ref{tab:capvqa_supp}.
The results show that the original CLIP-based LLaVA almost completely fails under such strong attacks, indicating its severe vulnerability to adversarial perturbations.
Among adversarially fine-tuned baselines, FARE \cite{schlarmann2024robust} still maintains relatively strong captioning performance, while TeCoA \cite{mao2022understanding} and PMG-AFT \cite{wang2024pre} achieve competitive results on some VQA datasets.
However, DAFT consistently achieves the best overall robustness under this stronger attack setting.
Specifically, compared with TeCoA, DAFT improves the average caption CIDEr score by 8.0 and the average VQA accuracy by 0.6\%.
Compared with FARE, DAFT further improves the average caption CIDEr score by 1.8 and the average VQA accuracy by 1.4\%.
Although TeCoA obtains the best result on VizWiz, DAFT achieves the highest performance on most datasets and obtains the best average performance across both captioning and VQA tasks.
These results further demonstrate that DAFT provides stronger adversarial generalization under more severe perturbations, complementing the main results in Sec. \ref{vqa-cap}.

\begin{table*}[ht]
\centering
\caption{Clean and Robust Performance of LLaVA with different vision encoders on different captioning and VQA datasets.
For caption datasets Flickr30 and COCO, we report their CIDEr scores. For other four VQA datasets, we report their VQA accuracy.
The highest accuracy for each dataset under different attack budgets is highlighted in bold.
Overall, DAFT demonstrates the best robust performance compared to other methods while achieving comparable clean performance.}
\label{tab:capvqa_supp}
\begin{adjustbox}{width=0.98\textwidth,keepaspectratio}
\begin{tabular}{c|l|cccccc|c|c}
\hline
\multicolumn{1}{l|}{Perturbation Budget} & Model                                     & Flickr30                              & COCO                                  & VQAv2                                 & OKVQA                                 & VizWiz                                & TextVQA                               & Average-Caption                           & Average-VQA                           \\ \hline
                                         & LLaVA-CLIP                                & 83.2                                  & 121.6                                 & 76.7                                  & 60.6                                  & 38.8                                  & 45.3                                  & 102.4                                 & 55.3                                  \\
                                         & LLaVA-TeCOA \cite{mao2022understanding}                              & 38.6                                  & 90.9                                  & 61.4                                  & 51.2                                  & 42.5                                  & 17.0                                  & 64.7                                  & 43.0                                  \\
                                         & LLaVA-PMG-AFT \cite{wang2024pre}                            & 40.4                                  & 91.8                                  & 60.3                                  & 50.7                                  & 42.4                                  & 16.7                                  & 66.1                                  & 42.5                                  \\
                                         & LLaVA-FARE  \cite{schlarmann2024robust}                              & 57.7                                  & 107.8                                 & 67.6                                  & 55.9                                  & 42.6                                  & 26.4                                  & 82.7                                  & 48.1                                  \\
\multirow{-5}{*}{clean}                  & \cellcolor[HTML]{EFEFEF}LLaVA-DAFT (ours) & \cellcolor[HTML]{EFEFEF}56.3          & \cellcolor[HTML]{EFEFEF}105.3         & \cellcolor[HTML]{EFEFEF}66.7          & \cellcolor[HTML]{EFEFEF}55.8          & \cellcolor[HTML]{EFEFEF}42.4          & \cellcolor[HTML]{EFEFEF}25.2          & \cellcolor[HTML]{EFEFEF}80.8          & \cellcolor[HTML]{EFEFEF}47.5          \\ \hline
                                         & LLaVA-CLIP                                & 3.9                                   & 7.9                                   & 7.7                                   & 1.6                                   & 0.1                                   & 0.8                                   & 5.9                                   & 2.5                                   \\
                                         & LLaVA-TeCOA \cite{mao2022understanding}                              & 12.0                                  & 41.3                                  & 38.3                                  & 28.5                                  & 29.0                                  & 8.5                                   & 26.6                                  & 26.1                                  \\
                                         & LLaVA-PMG-AFT  \cite{wang2024pre}                           & 11.9                                  & 40.6                                  & 36.6                                  & 29.1                                  & 29.2                                  & 8.7                                   & 26.3                                  & 25.9                                  \\
                                         & LLaVA-FARE  \cite{schlarmann2024robust}                              & 21.4                                  & 56.9                                  & 40.2                                  & 32.8                                  & 30.2                                  & \textbf{12.6}                         & 39.2                                  & 29.0                                  \\
\multirow{-5}{*}{$\epsilon$=2/255}                & \cellcolor[HTML]{EFEFEF}LLaVA-DAFT (ours) & \cellcolor[HTML]{EFEFEF}\textbf{23.0} & \cellcolor[HTML]{EFEFEF}\textbf{59.9} & \cellcolor[HTML]{EFEFEF}\textbf{41.4} & \cellcolor[HTML]{EFEFEF}\textbf{34.9} & \cellcolor[HTML]{EFEFEF}\textbf{31.4} & \cellcolor[HTML]{EFEFEF}\textbf{12.6} & \cellcolor[HTML]{EFEFEF}\textbf{41.5} & \cellcolor[HTML]{EFEFEF}\textbf{30.1} \\ \hline
                                         & LLaVA-CLIP                                & 2.1                                   & 4.1                                   & 1.5                                   & 0.1                                   & 0.1                                   & 0.0                                   & 3.1                                   & 0.4                                   \\
                                         & LLaVA-TeCOA  \cite{mao2022understanding}                             & 8.8                                   & 32.9                                  & 30.7                                  & 23.4                                  & 25.0                                  & 6.2                                   & 20.9                                  & 21.3                                  \\
                                         & LLaVA-PMG-AFT  \cite{wang2024pre}                           & 8.9                                   & 31.1                                  & 29.7                                  & 23.1                                  & 24.8                                  & 5.9                                   & 20.0                                  & 20.9                                  \\
                                         & LLaVA-FARE  \cite{schlarmann2024robust}                              & 16.1                                  & 45.2                                  & 33.0                                  & 26.0                                  & 25.9                                  & \textbf{9.4}                          & 30.6                                  & 23.6                                  \\
\multirow{-5}{*}{$\epsilon$=4/255}                & \cellcolor[HTML]{EFEFEF}LLaVA-DAFT (ours) & \cellcolor[HTML]{EFEFEF}\textbf{17.3} & \cellcolor[HTML]{EFEFEF}\textbf{48.1} & \cellcolor[HTML]{EFEFEF}\textbf{34.5} & \cellcolor[HTML]{EFEFEF}\textbf{27.9} & \cellcolor[HTML]{EFEFEF}\textbf{27.0} & \cellcolor[HTML]{EFEFEF}8.7           & \cellcolor[HTML]{EFEFEF}\textbf{32.7} & \cellcolor[HTML]{EFEFEF}\textbf{24.5} \\ \hline
                                         & LLaVA-CLIP                                & 0.8                                   & 2.1                                   & 0.0                                   & 0.0                                   & 0.0                                   & 0.0                                   & 1.4                                   & 0.0                                   \\
                                         & LLaVA-TeCOA \cite{mao2022understanding}   & 5.3                                   & 19.9                                  & 20.9                                  & 15.9                                  & \textbf{22.7}                         & 4.4                                   & 12.6                                  & 16.0                                  \\
                                         & LLaVA-PMG-AFT \cite{wang2024pre}          & 5.6                                   & 20.9                                  & 21.9                                  & 15.6                                  & 22.1                                  & 4.8                                   & 13.2                                  & 16.1                                  \\
                                         & LLaVA-FARE \cite{schlarmann2024robust}    & 9.3                                   & 28.3                                  & 22.4                                  & 16.1                                  & 17.0                                  & 5.3                                   & 18.8                                  & 15.2                                  \\
\multirow{-5}{*}{$\epsilon$=8/255}       & \cellcolor[HTML]{EFEFEF}LLaVA-DAFT (ours) & \cellcolor[HTML]{EFEFEF}\textbf{10.7} & \cellcolor[HTML]{EFEFEF}\textbf{30.6} & \cellcolor[HTML]{EFEFEF}\textbf{24.4} & \cellcolor[HTML]{EFEFEF}\textbf{16.5} & \cellcolor[HTML]{EFEFEF}19.7          & \cellcolor[HTML]{EFEFEF}\textbf{5.8}  & \cellcolor[HTML]{EFEFEF}\textbf{20.6} & \cellcolor[HTML]{EFEFEF}\textbf{16.6} \\ \hline
\end{tabular}
\end{adjustbox}
\end{table*}

\section{Performance on Different LVLMs}
\label{vqa-cap-OF}

\begin{table*}[ht]
\centering
\caption{Robust Performance of OpenFlamingo with different vision encoders on various datasets. 
For caption datasets Flickr30 and COCO, we report their CIDEr scores. For other four VQA datasets, we report their VQA accuracy.
The highest accuracy for each dataset under two attack budgets is highlighted in bold. Overall, OpenFlamingo-DAFT demonstrates strong performance under both clean and adversarial conditions.}
\label{tab:openflamingo}
\begin{adjustbox}{width=0.98\textwidth,keepaspectratio}
\begin{tabular}{c|l|cccccc|c|c}
\hline
\multicolumn{1}{l|}{Perturbation Budget} & Model                                     & Flickr30                              & COCO                                  & VQAv2                                 & OKVQA                                 & VizWiz                                & TextVQA                               & Average-Caption                           & Average-VQA                           \\ \hline
                                         & OpenFlamingo-CLIP                         & 0.5                                   & 1.7                                   & 0.4                                   & 1.1                                   & 0.9                                   & 0.0                                   & 1.1                                   & 0.6                                   \\
                                         & OpenFlamingo-TeCOA \cite{mao2022understanding}                        & 9.2                                   & 37.7                                  & 26.8                                  & 16.9                                  & 17.2                                  & 4.4                                   & 23.5                                  & 16.3                                  \\
                                         & OpenFlamingo-PMG-AFT \cite{wang2024pre}                      & 9.5                                   & 38.2                                  & 27.3                                  & 17.4                                  & 17.5                                  & 4.7                                   & 23.9                                  & 16.7                                  \\
                                         & OpenFlamingo-FARE \cite{schlarmann2024robust}                        & 15.3                                  & 47.9                                  & 28.1                                  & 22.1                                  & 19.7                                  & \textbf{7.8}                                   & 31.6                                  & 19.4                                  \\
\multirow{-5}{*}{e=2/255}                  & \cellcolor[HTML]{EFEFEF}OpenFlamingo-DAFT (ours) & \cellcolor[HTML]{EFEFEF}\textbf{16.6}         & \cellcolor[HTML]{EFEFEF}\textbf{51.3}          & \cellcolor[HTML]{EFEFEF}\textbf{29.6}          & \cellcolor[HTML]{EFEFEF}\textbf{23.4}          & \cellcolor[HTML]{EFEFEF}\textbf{21.7}          & \cellcolor[HTML]{EFEFEF}6.4           & \cellcolor[HTML]{EFEFEF}\textbf{34.0}          & \cellcolor[HTML]{EFEFEF}\textbf{20.3}          \\ \hline
                                         & OpenFlamingo-CLIP                         & 0.3                                   & 1.4                                   & 0.2                                   & 0.0                                   & 0.0                                   & 0.0                                   & 0.9                                   & 0.1                                   \\
                                         & OpenFlamingo-TeCOA \cite{mao2022understanding}                        & 5.5                                   & 26.7                                  & 23.2                                  & 14.7                                  & 15.9                                  & 2.8                                   & 16.1                                  & 14.2                                  \\
                                         & OpenFlamingo-PMG-AFT \cite{wang2024pre}                     & 5.7                                   & 27.1                                  & 23.7                                  & 14.5                                  & 16.4                                  & 3.1                                   & 16.4                                  & 14.4                                  \\
                                         & OpenFlamingo-FARE \cite{schlarmann2024robust}                         & 9.1                                   & 38.4                                  & 25.9                                  & 18.1                                  & 17.2                                  & \textbf{4.6}                                   & 23.8                                  & 16.5                                  \\
\multirow{-5}{*}{e=4/255}                  & \cellcolor[HTML]{EFEFEF}OpenFlamingo-DAFT (ours) & \cellcolor[HTML]{EFEFEF}\textbf{10.2}         & \cellcolor[HTML]{EFEFEF}\textbf{40.7}          & \cellcolor[HTML]{EFEFEF}\textbf{26.6}          & \cellcolor[HTML]{EFEFEF}\textbf{19.5}          & \cellcolor[HTML]{EFEFEF}\textbf{18.8}          & \cellcolor[HTML]{EFEFEF}4.4           & \cellcolor[HTML]{EFEFEF}\textbf{25.5}          & \cellcolor[HTML]{EFEFEF}\textbf{17.3}          \\ \hline
\end{tabular}
\end{adjustbox}
\end{table*}

We also conduct experiments on OpenFlamingo \cite{awadalla2023openflamingo}, a model which uses the same CLIP visual encoder as its visual backbone. 
The attack setup is identical to the one in Sec. \ref{vqa-cap}. 
Tab. \ref{tab:openflamingo}  presents the evaluation results on OpenFlamingo using different adversarial fine-tuning methods. 
The following analysis mainly focuses on the robust performance of the OpenFlamingo model and the transferability of the DAFT method.
Tab. \ref{tab:openflamingo} shows that OpenFlamingo demonstrates overall weaker robustness against adversarial attacks. 
On various caption datasets and VQA datasets, its robust performance is poorer compared to LLaVA. 
However, OpenFlamingo-DAFT shows a significant performance improvement under different adversarial perturbations compared to the original OpenFlamingo model (e.g., OpenFlamingo-CLIP), proving the effectiveness of the DAFT method in enhancing robustness. 
A comparison between OpenFlamingo-DAFT and LLaVA-DAFT indicates that while both benefit from the DAFT method, OpenFlamingo exhibits lower robust performance on certain datasets (e.g., COCO), not reaching the performance level of LLaVA. 
However, OpenFlamingo-DAFT demonstrates a remarkable improvement under adversarial attacks, especially on more challenging datasets such as TextVQA and OKVQA, where its robustness significantly outperforms other fine-tuning methods. 
The comparison between LLaVA and OpenFlamingo highlights the transferability of the DAFT method. 
Despite architectural differences between the two models, DAFT enhances robustness on both models. 
This suggests that DAFT is not reliant on a specific visual encoder or model architecture, and its ability to improve robustness transfers across different models.

\section{Black-box Transferability Evaluation}
\label{sec:blackbox_transfer}

\begin{table}[t]
\centering
\caption{Black-box transfer evaluation between LLaVA and OpenFlamingo under full-model attacks with $\epsilon$ = 4/255.
All adversarially fine-tuned vision encoders are trained with the perturbation budget of $\epsilon$ = 4/255.
``LLaVA $\rightarrow$ OF'' denotes adversarial examples generated by attacking the full LLaVA model and evaluated on OpenFlamingo, while ``OF $\rightarrow$ LLaVA'' denotes the reverse transfer direction.
For captioning tasks, we report the average CIDEr score; for VQA tasks, we report the average VQA accuracy.}
\label{tab:blackbox_transfer}
\begin{adjustbox}{width=\linewidth,keepaspectratio}
\begin{tabular}{l|cc|cc}
\hline
\multirow{2}{*}{Model} 
& \multicolumn{2}{c|}{Caption} 
& \multicolumn{2}{c}{VQA} \\ 
\cline{2-5}
& LLaVA $\rightarrow$ OF 
& OF $\rightarrow$ LLaVA 
& LLaVA $\rightarrow$ OF 
& OF $\rightarrow$ LLaVA \\ 
\hline
CLIP                                      & 10.2                                  & 29.6                                  & 17.7                                  & 43.0                                  \\
TeCoA \cite{mao2022understanding}         & 22.6                                  & 55.0                                  & 22.0                                  & 43.6                                  \\
PMG-AFT \cite{wang2024pre}                & 34.3                                  & 57.5                                  & 23.1                                  & 44.7                                  \\
FARE \cite{schlarmann2024robust}          & 35.5                                  & 64.8                                  & 23.8                                  & 47.5                                  \\
\cellcolor[HTML]{EFEFEF}DAFT (ours)       & \cellcolor[HTML]{EFEFEF}\textbf{39.5} & \cellcolor[HTML]{EFEFEF}\textbf{65.4} & \cellcolor[HTML]{EFEFEF}\textbf{24.8} & \cellcolor[HTML]{EFEFEF}\textbf{48.1} \\
\hline
\end{tabular}
\end{adjustbox}
\end{table}

We further evaluate the black-box transferability of adversarial examples between LLaVA and OpenFlamingo.
The evaluated tasks are consistent with those in Sec.~\ref{vqa-cap}, including image captioning and VQA.
Different from the previous white-box evaluation, where adversarial examples are generated and evaluated on the same LVLM, we perform full-model attacks on one LVLM and directly evaluate the generated adversarial examples on the other LVLM.
Specifically, we consider two transfer directions: attacking LLaVA and evaluating on OpenFlamingo, denoted as LLaVA $\rightarrow$ OF, and attacking OpenFlamingo and evaluating on LLaVA, denoted as OF $\rightarrow$ LLaVA.
All adversarially fine-tuned vision encoders used in this experiment are trained under the perturbation budget of $\epsilon$ = 4/255, and the black-box transfer attacks are also conducted with $\epsilon$ = 4/255.
This setting evaluates whether the robustness improvement brought by different adversarial fine-tuning methods can generalize across LVLM architectures under transferred black-box attacks.

We do not include zero-shot classification in this black-box transfer evaluation because the classification setting only attacks the standalone CLIP model.
Unlike captioning and VQA tasks, zero-shot classification does not involve different LVLM architectures or full-model generation pipelines.
Therefore, it does not directly reflect the cross-model transferability problem considered here.

Tab.~\ref{tab:blackbox_transfer} shows that DAFT consistently achieves the best performance across both transfer directions and both task types.
On captioning tasks, DAFT outperforms FARE by 4.0 CIDEr points under the LLaVA $\rightarrow$ OF setting and by 0.6 CIDEr points under the OF $\rightarrow$ LLaVA setting.
On VQA tasks, DAFT further improves the average accuracy over FARE by 1.0\% and 0.6\% under the two transfer directions, respectively.
Compared with category-label supervised methods such as TeCoA \cite{mao2022understanding} and PMG-AFT \cite{wang2024pre}, DAFT also shows more stable robustness across different transfer settings.
These results indicate that DAFT not only improves white-box adversarial robustness, but also provides better robustness against transferred black-box attacks.
Moreover, since the adversarial perturbations are generated by attacking the full LVLM rather than only the vision encoder, the results further demonstrate that the robustness learned by DAFT can better generalize to practical cross-model attack scenarios.

\end{document}